\documentclass{article}
\usepackage{spconf,amsmath,epsfig}

\title{Joint multi-modal Self-Supervised pre-training in Remote Sensing: Application to Methane Source Classification}
\name{Paul Berg, Minh-Tan Pham, Nicolas Courty\thanks{This work is funded by the ANR AI chair OTTOPIA under reference ANR-20-CHIA-0030. This work was performed using HPC resources from GENCI-IDRIS (Grant 2022-AD011013514).}}
\address{IRISA, Université de Bretagne Sud, UMR 6074, F-56000 Vannes, France}
\begin{document}
%
\maketitle

\begin{abstract}

With the current ubiquity of deep learning methods to solve computer vision and remote sensing specific tasks, the need for labelled data is growing constantly. However, in many cases, the annotation process can be long and tedious depending on the expertise needed to perform reliable annotations. In order to alleviate this need for annotations, several self-supervised methods have recently been proposed in the literature. The core principle behind these methods is to learn an image encoder using solely unlabelled data samples. In earth observation, there are opportunities to exploit domain-specific remote sensing image data in order to improve these methods. Specifically, by leveraging the geographical position associated with each image, it is possible to cross reference a location captured from multiple sensors, leading to multiple views of the same locations. In this paper, we briefly review the core principles behind so-called joint-embeddings methods and investigate the usage of multiple remote sensing modalities in self-supervised pre-training. We evaluate the final performance of the resulting encoders on the task of methane source classification.

\end{abstract}


\section{Introduction}
\label{sec:intro}

By considering the large amount of remote sensing data collected daily by different satellite sensors, there is a growing interest in leveraging those large-scale data to perform downstream tasks such as scene classification. Deep learning methods have imposed themselves as the de-facto tools for most tasks operating on geospatial data~\cite{cheng2020remote}. However, supervised neural networks require large quantities of annotations in order to perform at full capabilities. Therefore, recent methods based on self-supervised learning (SSL) have been proposed to reduce the need for labels for training in computer vision~\cite{liu2021self} as well as in remote sensing applications~\cite{berg2022self,wang2022sslrs}. Among these methods, joint-embeddings SSL frameworks have shown impressive results, even learning image representations on par with or better than state of the art supervised models~\cite{caron2020unsupervised}. In remote sensing, early works have been proposed toward learning across modalities in a self-supervised manner by exploiting the multi-modal nature of various remote sensing datasets~\cite{scheibenreif2022contrastive,wang2022dinomm,jain2022self}. Indeed, the geographical locations associated with each samples can allow the creation of pairs of samples capturing the same geographical location. These pairs can then be seen as different views of the same location and incorporated in a self-supervised learning pipeline. In this paper, we build upon these works which focus on learning on only two modalities to develop a framework for self-supervised pre-training across several image modalities. We work with up to three but there could be an arbitrary number of modalities if available. The proposed method is then tested on the task of methane source classification using the Meter-ML dataset~\cite{zhu2022meter}. Our experiments show that scaling the number of modalities in SSL pre-training can improve the performance on downstream tasks. Interestingly, these results remains true when the downstream task inputs only a single modality, highlighting the potential of leveraging new modalities in self-supervised learning applied to remote sensing tasks. We also investigate the impact of artificial augmentations in this pre-training pipeline compared to using only the geographical cross references.

\section{Methodology}
\label{sec:methodo}

\begin{figure*}[htb!]
    \centering
    \includegraphics[width=0.85\textwidth]{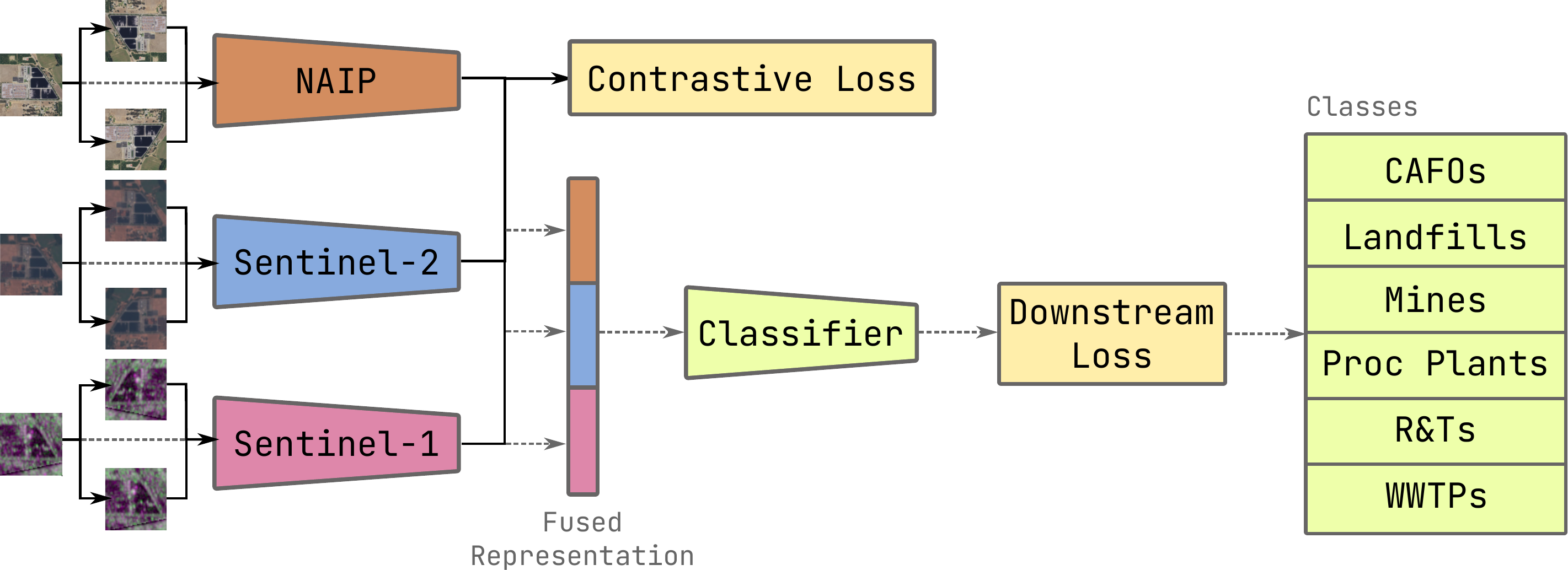}
    \caption{Architecture of our multi-modal pre-training and finetuning processes. Black arrows represent the forward data flow during pre-training while dashed gray arrows represent the forward data flow during finetuning. To ease the reading, we illustrate a specific use-case of methane source classification using the from the Meter-ML dataset~\cite{zhu2022meter} with three modalities: NAIP, Sentinel-2 and Sentinel-1 images. In the downstream classification task, methane emitting sources are divided in six classes.}
    \label{fig:architecture}
\end{figure*}

Joint-embedding methods enable the pre-training of models in a self-supervised manner by using multiple views from the same data point~\cite{chen2020simple,chen2020improved,grill2020bootstrap,chen2021exploring}. In most cases where the unlabeled data used for pre-training come from a single modality, a common technique is to create different views based on randomly artificial augmentations such as cropping, color jittering, rotations. Given a sample view $x$, we call views from the same sample positives $x^+$ and we refer to views from a different sample as negatives $x^-$. The main objective of joint-embedding methods is to maximize the alignment between the representation of a view and representations of its corresponding positives. However, this objective is not sufficient since the model can trivially converge to a constant representation minimizing the distance between all representations and producing un-discriminative representations (i.e., collapse problem). To circumvent this problem, two families of solutions have been proposed. Firstly, contrastive methods~\cite{chen2020simple,he2020momentum} leverage other samples in the batch as so-called negatives so that the learning objective not only minimizes the distance of each sample to its positive views but also maximizes its distance to negatives, thus preventing the latent space from collapsing in a single point. Secondly, several regularization-based or architectural methods rely on either additional loss terms~\cite{zbontar2021barlow,bardes2022vicreg} or adding network asymmetry~\cite{grill2020bootstrap,chen2021exploring} for the different augmented views. In this paper, we focus on contrastive methods due to their popularity in SSL. Also, they have been until now the most widely used SSL approaches in remote sensing \cite{berg2022self}. The most common version of the contrastive loss is called InfoNCE~\cite{chen2020simple}. It maximizes the alignment between positive views from the same sample while minimizing the alignment with views from other samples present in the same batch using a softmax cross-entropy. Given an encoder model $f_\theta(\cdot)$ parameterized by $\theta$, the loss for a single view is:

\begin{equation}
    \mathcal L_\text{NCE}(x;\theta) = -\sum_{\hat x^+\in x^+} \frac{
        \exp(\langle f_\theta(x), f_\theta(\hat x^+)\rangle / \tau)
    }{\sum_{\hat x\in\Omega(x)} \exp(\langle f_\theta(x), f_\theta(\hat x)\rangle / \tau)}
    \label{eq:contrastive_single}
\end{equation}

where $\tau > 0$ is a temperature parameter used to control the sharpness of the distribution generated by the softmax operator and $\langle x, y\rangle$ refers to the dot product between $x$ and $y$ both normalized to unit vectors, as the cosine similarity. $\Omega(x)$ refers to the set of augmented views present in the batch excluding $x$, namely $\Omega(x) = x^+ \cup x^-$. During self-supervised pre-training, the loss is applied to every generated view. Therefore, for a batch $x$ of original $N$ samples, each augmented with $T$ views, the following loss function will be computed:

\begin{equation}
    \mathcal L_\theta = \frac 1{T\times N} \sum^{T\times N}_{i = 1} \mathcal L_\text{NCE}(x_i;\theta).
\end{equation}

As each modality contains a different number of channels, we pre-train a backbone for each one. These backbones can therefore be used on their own as a discriminative initialization after self-supervised pre-training. During the downstream tasks, a single feature can be used for a geographical location by fusing representations from each input modalities.

\section{Experimental Study}
\label{sec:expe}

\begin{table*}[htb!]
    \centering
    \begin{tabular}{l|ccccccc}
          Pre-training & \multicolumn{7}{c}{Downstream} \\
                        & S1 & S2 & NAIP & S1 + S2 &  S1 + NAIP & S2 + NAIP & S1 + S2 + NAIP \\
         \hline
         None           & 47.37\% & 64.29\% & 62.03\% & 65.04\% & 63.16\% & 68.42\% &  65.79\% \\
         \hline
         S1             & 51.13\% & -       & -       & -       & -                & -       & - \\
         S2             & -       & 70.30\% & -       & -       & -                & -       & - \\
         NAIP           & -       & -       & 66.92\% & -       & -                & -       & - \\
         S1 + S2        & 53.76\% & 71.80\% & -       & 71.80\% & -                & -       & - \\
         S1 + NAIP      & 56.39\% & -       & 70.68\% & -       & \textbf{72.18\%} & -       & - \\
         S2 + NAIP      & -       & 71.43\% & 68.42\% & -       & -                & 72.18\% & - \\
         S1 + S2 + NAIP & \textbf{58.65\%} & \textbf{72.56\%} & \textbf{72.93\%}  & \textbf{72.18\%} & 68.80\% & \textbf{73.31\%} & \textbf{73.68\%}  \\
         \hline
         S1 + S2       $^\star$ & 50.00\% & 69.55\% & -       & 65.04\% & -       & -       & -        \\
         S1 + NAIP     $^\star$ & 55.26\% & -       & 70.30\% & -       & 60.90\% & -       & -        \\
         S2 + NAIP     $^\star$ & -       & \textbf{72.56\%} & 69.92\% & -       & -       & 72.93\% & -        \\
         S1 + S2 + NAIP$^\star$ & 52.63\% & 71.05\% & 69.92\% & 65.41\% & 65.79\% & 69.92\%  & 69.17\%  \\
    \end{tabular}
    \caption{Accuracy using different pre-training and finetuning scenarios. When pre-training is None, it refers to randomly initialized baseline models. Pre-trainings annotated with $^\star$ denotes using no random augmentations and using only view (the original ) by modality during pre-training. In this setting, the self-supervised pre-training cannot be done on a single modality.}
    \label{tab:results}
\end{table*}

The Meter-ML~\cite{zhu2022meter} dataset used in our experiments for methane source classification contains multiple modalities for each geographic coordinates. Each methane-emitting present facility in the dataset therefore has corresponding Sentinel-1, Sentinel-2 and NAIP sensor captures. The dataset contains facilities from six different classes:  concentrated animal feed operations (CAFOs), coal mines, landfills, natural gas processing plants (Proc Plants), oil refineries and petroleum terminals (R\&Ts), and wastewater treatment plants (WWTPs). To experiment with both optical and SAR data as well as both low and high resolution, we pick sensor views from Sentinel-1 (VH and VV) and Sentinel-2 (RGB and NIR) at 10-m resolution as well as NAIP (RGB and NIR) at 1-m resolution. The proposed architecture is composed of a backbone for each modality used during pre-training (see Figure~\ref{fig:architecture}). We use an AlexNet~\cite{krizhevsky2017imagenet} for Sentinel-1 (S1) and Sentinel-2 (S2) and a ResNet18~\cite{he2016deep} model for NAIP. To evluate the impact of artificial augmentations, we compare self-supervised pre-training with artificial augmentations and without. When artificial augmentations are used, we generate two versions of the same image for each modality using random augmentations. Our set of augmentations includes random horizontal and vertical flips and a random cropping with a conservative resized crop with a scale of at least 90\% of the original image. With artificial augmentations, it results in each augmented view having a randomly augmented positive in the same modality as well as multiple augmented positives in other available modalities whereas without artificial augmentations, the view from a modality only has corresponding positives in other modalities. The Negative class samples present in the Meter-ML dataset are only used as negatives in the contrastive loss (see equation~\ref{eq:contrastive_single}). Models are pre-trained for 120 epochs. When using multiple modalities during finetuning, representations from each backbone are concatenated to produce a single feature vector for the sample which is then fed to the classifier. For methane source classification, we finetune models for 100 epochs. The training and validation sets are set following the official split of the Meter-ML dataset. Results can be seen on Table~\ref{tab:results}.

From the table, self-supervised pre-training consistently improves the performance compared to randomly initialized models. The multi-backbone architecture also scales with the number of modalities even when certain modalities are removed for the downstream task. The best overall performance is obtained when combining all modalities for the downstream classification task. Each modality therefore contains information relevant to the classification task that the methane source classifier is able to exploit. It is interesting to highlight that using fusion only during the downstream task and with a random initialization leads to a worst performance than using only Sentinel-2 data during pre-training and finetuning. This means that this modality could provide the most important information for classification and that self-supervised pre-training allows a discriminative initialization, which gives a better performance on the scene classification downstream task.

Our experiments without artificial augmentations show interesting results. Mainly, when pre-training with modalities which are different in nature like SAR and optical data (S1+S2, S1+NAIP), the results are better with artificial augmentations. This phenomenon suggests that the single positive from the other modality is hard to align to. Indeed, adding artificial augmentations improves the pre-training performance by also offering in-modality positives. With only optical pairs from different modalities (when pre-training with Sentinel-2 and NAIP for example), the drop in performance is less severe. Therefore in this case, artificial augmentations have less impact on the downstream performance. In any case, the pre-training performance with only a single modality and random augmentations performs worse than pre-training with multiple modalities, but remains better than random initialization for our chosen downstream task of scene classification with finetuning.

\section{Conclusion}
\label{sec:conclusion}

We performed evaluations on a multi-modal self-supervised pre-training pipeline. By leveraging the geographical pairs of sensor captures as multiple views, the performance of self-supervised pre-training can be improved compared to using a single modality pre-training with only artificial augmentations. In order to improve the performance of those pre-trained models, we leave to future work the evaluation of different fusing methods for the downstream task. Another interesting direction for improvement is the sharing of model weights between modalities. Indeed, our proposed approach requires having a entire set of encoder weights for each modality. Finally, we would like to explore in-depth the impact of the Negative class on the final performance of the model and how other unrelated datasets can be used as Negatives only during the pre-training phase. Overall, multi-modal self-supervised learning provides a better initialization than single modal self-supervised learning for methane source classification. Future work could also investigate whether or not these results generalize to the more general downstream task of remote sensing scene classification. Hopefully, this shows the interest of using other datasets during pre-training than only the ones concerned by the downstream tasks. There are also other opportunities to include remote sensing specific data in the pre-training phase in the form of domain specific augmentations which should continue to be investigated because these often have more impact than the specific type of self-supervised loss used.

\bibliographystyle{IEEEbib}
\bibliography{main}

\end{document}